\documentclass[letterpaper]{article}

\usepackage{aaai25}
\usepackage{times}
\usepackage{helvet}
\usepackage{courier}
\usepackage[hyphens]{url}
\usepackage{graphicx}
\usepackage{natbib}
\usepackage{caption}
\usepackage{multirow}
\usepackage{placeins}
\usepackage{graphicx}
\usepackage{soul}
\usepackage[linesnumbered,ruled,vlined]{algorithm2e}
\usepackage{algpseudocode}
\usepackage{amsmath}
\SetKwInput{KwInput}{Input}
\usepackage{lipsum}
\usepackage{amssymb}
\usepackage{seqsplit}
\usepackage{multicol}
\usepackage{adjustbox}
\usepackage{xcolor}
\usepackage{paralist}
\usepackage{float}
\newsavebox{\ebbsbox}
\usepackage{fancyhdr}
\let\oldnl\nl
\newcommand{\nonl}{\renewcommand{\nl}{\let\nl\oldnl}}

\title{
\begin{tabular}{cc}
\multirow{2}{*}{\includegraphics[height=1cm]{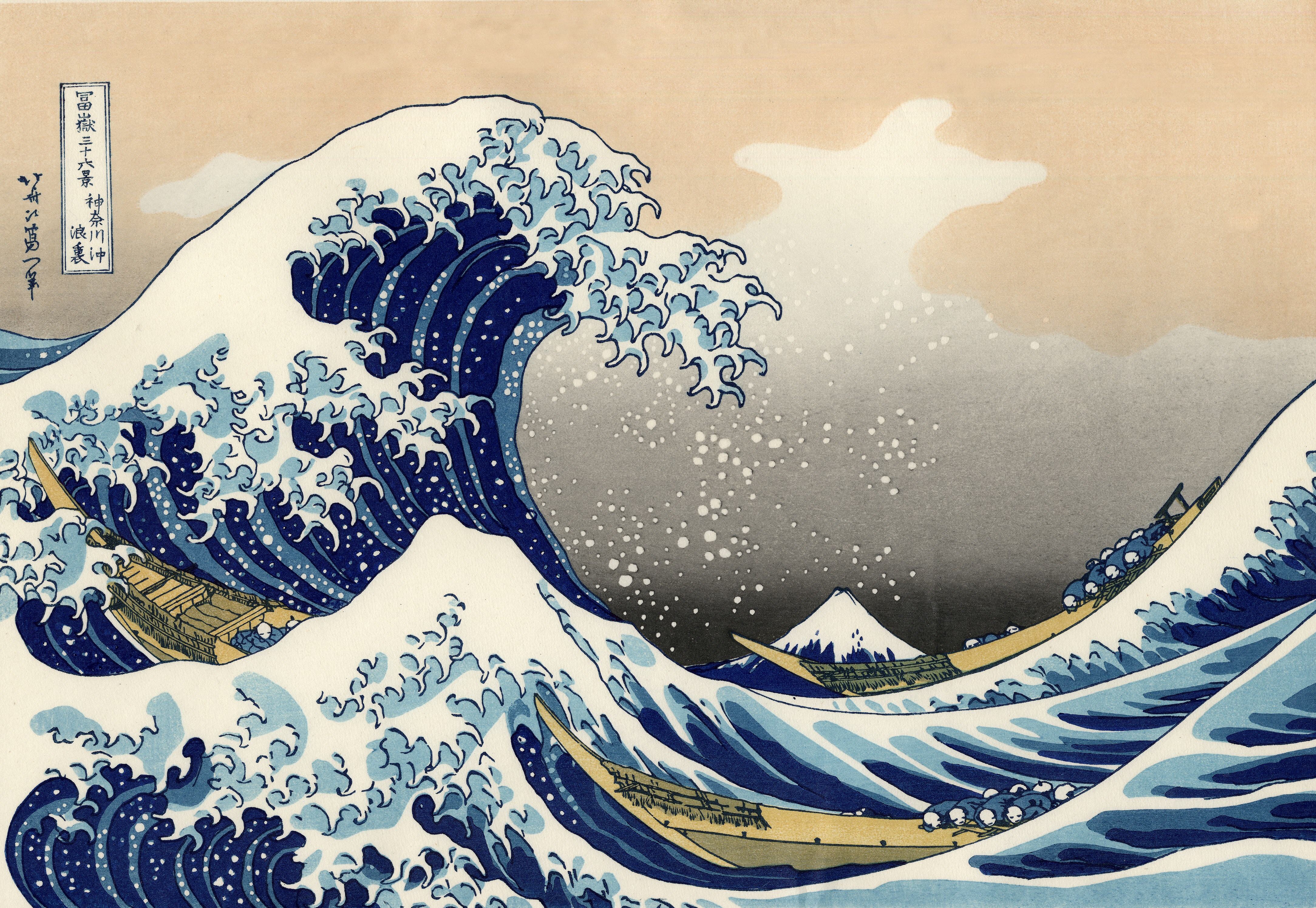}} & EBBS: An Ensemble with Bi-Level Beam Search \\
 & for Zero-Shot Machine Translation
\end{tabular}
}
\author{
    Yuqiao Wen$^{1,}$\thanks{Work partially done during Mitacs internship at RBC Borealis.}, Behzad Shayegh$^{1,*}$, Chenyang Huang$^1$, Yanshuai Cao$^2$, Lili Mou$^{1,3}$ \\
}
\affiliations{
    $^1$Dept. Computing Science, Alberta Machine Intelligence Institute (Amii), University of Alberta \\
    $^2$RBC Borealis \quad  $^3$Canada CIFAR AI Chair, Amii\\
    yq.when@gmail.com \quad the.shayegh@gmail.com \quad chenyangh@ualberta.ca \\
    yanshuai.cao@borealisai.com \quad doublepower.mou@gmail.com \\
}

\usepackage{bibentry}

\begin{document}

\maketitle

\begin{abstract}
The ability of zero-shot translation emerges when we train a multilingual model with certain translation directions; the model can then directly translate in unseen directions. Alternatively, zero-shot translation can be accomplished by pivoting through a third language (e.g., English). In our work, we observe that both direct and pivot translations are noisy and achieve less satisfactory performance. We propose EBBS, an ensemble method with a novel bi-level beam search algorithm, where each ensemble component explores its own prediction step by step at the lower level but all components are synchronized by a ``soft voting'' mechanism at the upper level. Results on two popular multilingual translation datasets show that EBBS consistently outperforms direct and pivot translations, as well as existing ensemble techniques. Further, we can distill the ensemble's knowledge back to the multilingual model to improve inference efficiency;
profoundly, our EBBS-distilled model can even outperform EBBS as it learns from the ensemble knowledge.
\end{abstract}

\begin{links}
    \link{GitHub}{https://github.com/MANGA-UOFA/EBBS}
\end{links}

\renewcommand{\headrulewidth}{0pt}
\cfoot{Accepted by AAAI 2025}
\rhead{}
\thispagestyle{fancy}

\section{Introduction} \label{sec:intro}

Machine translation is a widely applicable NLP task that aims to translate a text from a source language to a target language~\cite{brown-etal-1990-statistical,bahdanau2014neural}.
The Transformer architecture~\cite{vaswani2017-transformer} and pretrained large language models~\cite{radford2019language,lewis-etal-2020-bart} have largely improved translation performance, especially in the supervised setting~\cite{t52020}, where a model can learn from large volumes of parallel corpora.
However, machine translation remains challenging for low-resource languages, because there are not enough data for large neural networks to learn these languages~\cite{radford2019language,NEURIPS2023_9d89448b}.

We specifically focus on multilingual translation in the \textit{zero-shot} setting, where the system is required to translate between unseen language pairs.
Since collecting parallel data and training individual models for every translation pair are prohibitively expensive, it is common to build a single multilingual system~\cite{johnson-etal-2017-googles,JMLR:v22:20-1307} that can perform translation for all language pairs, most of which are zero-shot translation directions that do not involve a high-resource language (e.g., English).
These models work by prepending a language-indicator token; the zero-shot translation ability emerges as the model generalizes from trained language pairs and is able to perform \textit{direct translation} for unseen ones~\citep{liu-etal-2021-improving-zero,wicks-duh-2022-effects}. The main drawback of such multilingual models is that they are noisy in the zero-shot setting due to the lack of supervision, and as a result, they tend to generate low-quality translations~\citep{zhang-etal-2020-improving,liu-etal-2021-improving-zero}.

Alternatively, zero-shot translation can be performed by \textit{pivoting}~\citep{wu-wang-2007-pivot,wu-wang-2009-revisiting}, where the model first translates the input into a high-resource language such as English, which is then translated to the target language.
However, pivoting requires two translation steps, often leading to an accumulation of errors~\cite{babych-etal-2007-translating,gu-etal-2019-improved}.

In this paper, we propose an ensemble approach that aggregates direct and pivot translations in order to build a stronger multilingual translation model from weak ones. Building an ensemble for text generation is nuanced as it involves a sequence of word predictions.
Word-level ensembles aggregate predictions at each generation step, which is usually achieved by averaging the predicted probabilities~\cite{sennrich-etal-2016-edinburgh,Freitag2017EnsembleDF,vishnu-kudlu-shanbhogue-etal-2023-improving}.
This may not be ideal for zero-shot translation as the predictions are too noisy, making the averaged probabilities overly smooth.
On the other hand, minimum Bayes risk decoding (MBR)~\cite{MBRbook} can be considered a sequence-level voting ensemble, but existing MBR methods are only able to \textit{select} from weak and noisy candidates given by the direct and pivot translations.

To this end, we propose an ensemble decoding algorithm with bi-level beam search (EBBS). Our EBBS performs two levels of beam search at each generation step: at the lower level, beam search is applied individually to each ensemble component; at the upper level, the ensemble maintains a shared beam by voting and synchronizing the candidates (sub-sequences) in lower-level beams.
Unlike word-level ensembles~\cite{Freitag2017EnsembleDF,vishnu-kudlu-shanbhogue-etal-2023-improving}, EBBS does not average the predicted distributions, encouraging individual predictors to explore their own preferences;
unlike sequence-level MBR ensembles~\cite{kobayashi-2018-frustratingly,eikema-aziz-2020-map}, EBBS does not select from a candidate set, and thus is more flexible since votings are performed throughout the generation process.

We conducted experiments on IWSLT~\citep{cettolo-etal-2017-overview} and Europarl~\citep{koehn-2005-europarl}, two popular multilingual datasets for zero-shot machine translation.
Results show that EBBS can generate high-quality translations and outperform existing ensemble techniques.
In addition, we used EBBS-generated data for distillation to further improve the multilingual model.
The experiment shows that such a distilling process encourages the model to learn from high-quality translations produced by EBBS, allowing it to outperform EBBS with no inference overhead compared with direct translation.

\section{Related Work}

\textbf{Machine translation.} In NLP, machine translation is a long-standing task that aims to rewrite text from one language to another without changing the meaning.
Traditional research in translation has been mainly centered on the supervised setting, utilizing manually crafted rules~\cite{forcada2011apertium,dugast-etal-2007-statistical} and statistical methods~\cite{brown-etal-1990-statistical,koehn2009statistical};
more recently, neural machine translation systems have considerably improved the performance~\cite{vaswani2017-transformer,t52020}.
However, translation remains challenging for low-resource languages, where neural models do not have enough parallel data to train on.

Translation for low-resource languages largely relies on  \textit{zero-shot} techniques, where no parallel text is available for a particular translation direction. In general, zero-shot translation can be accomplished in a monolingual or multilingual setting.
With monolingual data, the most common approach is to build language-specific autoencoders that share the same latent space of semantics; translation is then achieved by plugging in the decoder of the desired language~\cite{lample2018unsupervised,lample-etal-2018-phrase,mohiuddin-joty-2020-unsupervised}.

In this paper, we focus on the multilingual setting, where one model can translate between multiple languages~\cite{10.1145/3406095}.
Usually, parallel texts only exist for a high-resource language such as English, leaving translations between low-resource languages zero-shot (e.g., Italian to Dutch)~\cite{johnson-etal-2017-googles,JMLR:v22:20-1307}.
In this setting, the most common approach is to train the multilingual model on English-centric data, and the zero-shot translation ability naturally emerges during the training process~\cite{johnson-etal-2017-googles,workshop2022bloom}.

A key challenge for multilingual models is task interference, where too many languages tend to degrade model performance~\cite{zaremoodi-etal-2018-adaptive,wang-etal-2020-negative}.
As a result, research in this direction has been alleviating such interference by developing various parameter-separation schemes~\cite{baziotis-etal-2022-multilingual,chronopoulou-etal-2023-language} and using gradient-based methods to update language-specific parameters~\cite{Wang_Zhang_2022,he-etal-2023-gradient}.
In our work, we use a standard Transformer model following~\citet{johnson-etal-2017-googles} and \citet{liu-etal-2021-improving-zero}. Our proposed ensemble algorithm EBBS is compatible with the above approaches, as it is agnostic to model architectures.

\textbf{Ensemble methods.} In a model ensemble, multiple machine learning systems are integrated so as to form a stronger one~\citep{dong2020survey,yang2023survey}.
\textit{Bagging}, a classic ensemble technique, works by training multiple models with different portions of data and combining their predictions through averaging or voting~\citep{breiman1996bagging,buhlmann2002analyzing}.
Another popular ensemble approach is \textit{boosting}, where different models are trained sequentially, with each subsequent model addressing the mistakes of the previous ones~\citep{schapire2003boosting,hastie2009multi,natekin2013gradient}.
Unfortunately, bagging and boosting are not compatible with our setting, because we build an ensemble with a single model.
Alternatively, \textit{stacking} combines the outputs by training a meta-model~\citep{WOLPERT1992241,GANAIE2022105151}, but this does not apply to our zero-shot setting either because we do not have groundtruth signals to train the meta-model. Even though these ensemble techniques may be directly applied to supervised generation~\citep{Freitag2017EnsembleDF,kobayashi-2018-frustratingly,hendy-etal-2021-ensembling}, they are not ideal as they do not take advantage of structured prediction. Our recent work has addressed the ensemble of tree structures~\citep{shayegh2023ensemble,shayegh-etal-2024-tree,shayegh-selection}, and in this paper we focus on text generation.

Unlike previous work, our EBBS performs bi-level beam search, exploring different components' own predictions and synchronizing them by a ``soft voting'' mechanism at every step. Our approach is specifically suited to the sequence generation process.

\section{Approach}

In this section, we first explain our ensemble components in \S\ref{subsec:components}.
In \S\ref{subsec:ebbs}, we propose EBBS, a novel ensemble decoding algorithm.
Finally, we describe in \S\ref{subsec:kd} knowledge distillation with EBBS-decoded outputs for efficiency considerations.

\subsection{Ensemble Components} \label{subsec:components}

In this work, we focus on zero-shot multilingual machine translation, which requires a system to perform translations for multiple languages, where some translation directions are unseen.

Specifically, our multilingual model is an encoder--decoder Transformer with a byte pair encoding tokenizer~\citep{sennrich-etal-2016-neural} shared among all languages.  The encoder can capture the semantics of tokens in different languages, whereas the decoder translates the encoded text into the desired language based on a target-language indicator token~\cite{johnson-etal-2017-googles,JMLR:v22:20-1307}.

We follow the standard English-centric training~\citep{johnson-etal-2017-googles,liu-etal-2021-improving-zero}, where the multilingual model is trained using parallel data with English on one side (e.g., German-to-English and English-to-Romanian). As mentioned in \S\ref{sec:intro}, the zero-shot ability emerges during such training, and the model is able to perform direct translation between unseen language pairs (e.g., German-to-Romanian)~\cite{10.1145/3406095,10.1145/3567592}. An alternative approach is pivot translation, where the multilingual model performs two translations using a high-resource language as a pivot (e.g., first translating German to English, and then English to Romanian).

However, both direct and pivot translations have major weaknesses: the quality of direct translation tends to be low due to the lack of parallel data, whereas pivot translation suffers from error accumulation as it requires two translation steps~\citep{babych-etal-2007-translating,gu-etal-2019-improved}.

In this paper, we would like to build an ensemble of direct and pivot translations to boost translation quality, where each translation path results in an ensemble component.
Commonly used ensemble methods such as averaging and voting may not work well for text generation. Voting, for example, chooses the most voted prediction, but in text generation, the components' votes often do not share anything in common, because there could be tens of thousands of tokens in the vocabulary. 
An averaging ensemble, on the other hand, averages the predicted distributions of all components, potentially leading to an overly smooth distribution.
Despite early success by~\citet{razmara-sarkar-2013-ensemble}, more recent studies report marginal or negative improvement for multi-pivot averaging ensemble~\citep{JMLR:v22:20-1307,gaikwad2024effective,mohammadshahi-etal-2024-investigating}.

\subsection{Our Proposed EBBS Algorithm} \label{subsec:ebbs}

\begin{figure}[t]
    \centering
    \includegraphics[width=1.0\columnwidth]{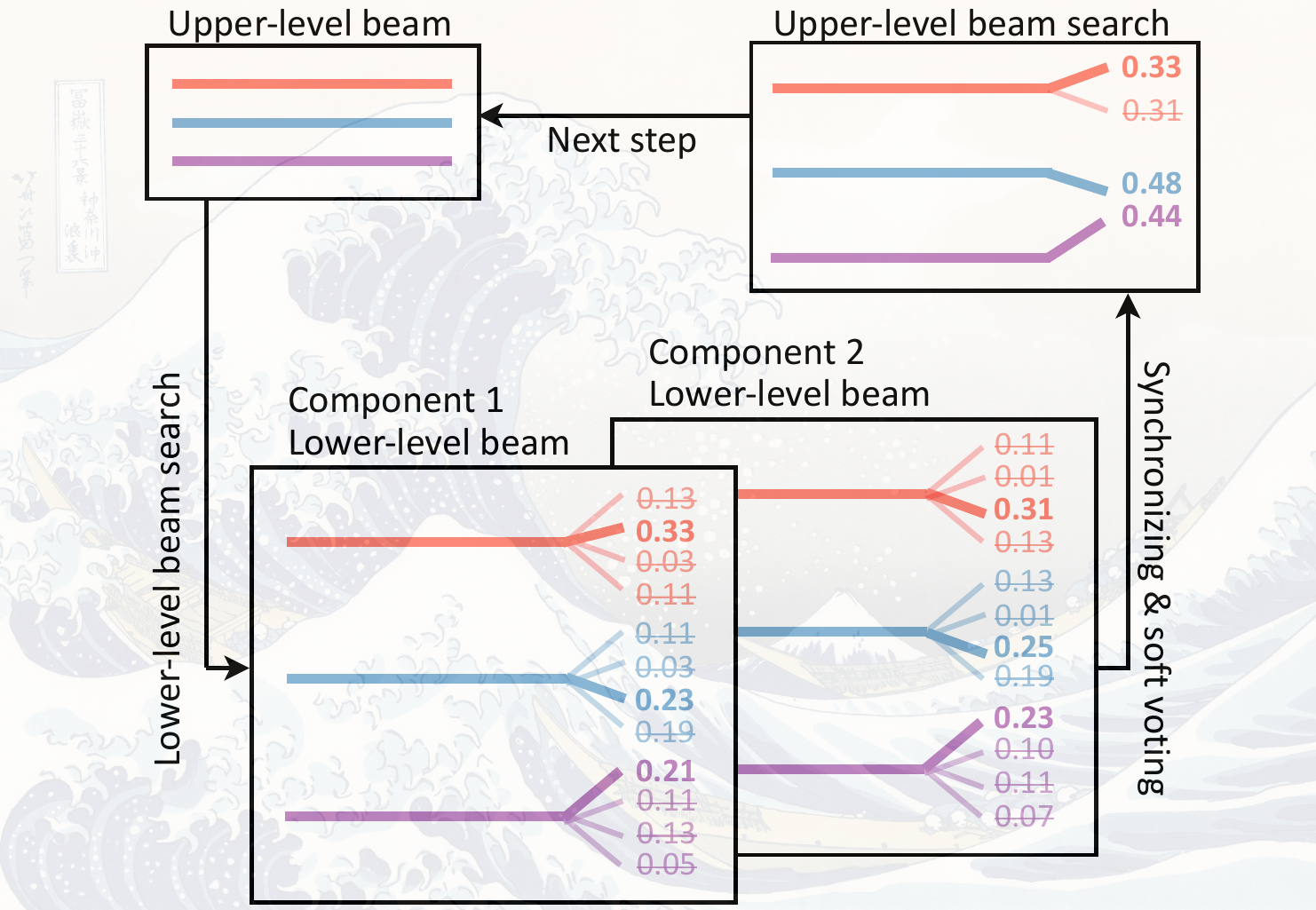}
    \caption{Illustration of our EBBS algorithm.}
    \label{fig:dynamics}
\end{figure}

We propose an ensemble with bi-level beam search (EBBS), a novel decoding algorithm that enables different ensemble components to collaborate and vote on each other's partial generations with two levels of beam search.

At the lower level, each ensemble component performs beam search individually, exploring its own preferred regions of the sentence space.
At the upper level, EBBS synchronizes the lower-level beam candidates through a voting mechanism, only keeping the most promising partial generations in a shared, upper-level beam.
This allows the ensemble components to vote out spurious partial candidates and improve zero-shot translation performance.

Concretely, we assume there are $K$ ensemble components $p_1, \cdots, p_K$, each predicting the probability of the next word given some prefix. 

For the $0$th decoding step, EBBS initializes the upper-level beam by $\overline{B}_0=\langle \textsc{bos},1\rangle$, suggesting that a sequence is forced to start with a special beginning-of-sequence token \textsc{bos} with probability $1$. 

For step $t$, each ensemble component performs lower-level beam search individually, based on the prefixes in the last step's shared beam  $\overline{B}_{t-1}$:
\begin{align}
     \underline{B}_{t,k} =  \operatorname{top-}\!Z \{\ & \langle \mathbf y_{1:t-1} \oplus \mathrm y,
    \ \  p \cdot p_k(\mathrm y|\mathbf y_{1:t-1}, \mathbf x) \rangle : \nonumber \\
    &
     \langle \mathbf y_{1:t-1}, p \rangle \in \overline{B}_{t-1}, \ \  \mathrm y \in V \ \}
\end{align}
for $k=1,\cdots, K$. Here, $\operatorname{top-}\!Z$ selects $Z$-many sequences with the highest probabilities, $\oplus$ represents string concatenation, $V$ is the vocabulary, and $p_k(\mathrm y|\mathbf y_{1:t-1}, \mathbf x)$ is the $k$th ensemble component's predicted probability at step $t$ given the prefix $\mathbf y_{1:t-1}$ and input~$\mathbf x$.

At the upper level, EBBS synchronizes the lower-level individual beams $\underline{B}_{t,k}$, for $k=1,\cdots, K$, into a shared,
upper-level beam through a soft-voting mechanism, where the candidate set $C_t$ is the union of the sequences in lower-level beams:
\begin{align}
    C_t =  \bigcup\nolimits_k \{ \mathbf y: \langle \mathbf y, p\rangle\in \underline{B}_{t,k}\}
\end{align}
We evaluate each candidate in $C_t$ and compute its overall vote as the sum of the probabilities. 
\begin{align}
    \overline{B}_t = \operatorname{top-}\!\!Z\! \scalebox{1.2}{$\Bigg\{$}
 \left\langle \mathbf y,\!\! \sum_{k:\ k=1,\cdots, K\atop
     \langle\mathbf y', p \rangle\in \underline{B}_{t,k}:\ \mathbf y'=\mathbf y } p \right\rangle :\  \mathbf y \in C_t\scalebox{1.2}{$\Bigg\}$}
    \label{eq:sum-vote}
\end{align}

In this way, the upper level synchronizes all ensemble components with the shared beam $\overline{B}_t$ for the next step of generation. 

Intuitively, our voting scheme gives an ensemble component $Z$-many votes, each weighted by the predicted probability. The votes (probabilities) are then tallied (summed) for each candidate to form the upper-level beam.
Our bi-level beam search terminates when we have $Z$-many terminated sequences in the shared beam, and returns the sequence with the highest score\footnote{For selecting the final output, we follow standard implementations and normalize the joint probability by length, i.e., taking the geometric mean of step-wise probabilities~\citep{wolf2019huggingface,ott-etal-2019-fairseq}. Otherwise, beam search algorithms are often biased towards short sequences~\cite{meister-etal-2020-beam}. 
} as the ensemble output. We provide the detailed pseudocode for EBBS in Algorithm~\ref{alg:ebbs} and an illustration in Figure~\ref{fig:dynamics}.

\textbf{Discussion.} Traditional beam search keeps a fixed-size beam of high-likelihood partial sequences. To build an ensemble with multiple predictors, it is tempting to directly average their probabilities
$p(\mathbf y | \mathbf x) = \frac1K \sum_{k=1}^K p_k(\mathbf y|\mathbf x)
$ as the score for beam search, which has been experimented in previous work~\cite{sennrich-etal-2016-edinburgh,vishnu-kudlu-shanbhogue-etal-2023-improving}.

However, our intuition suggests that such an approach may suffer from the \textit{over-smoothing problem}~\cite{wei2019neural,wen-etal-2023-f}: when multiple translations (known as \textit{modes}) are plausible given an input, the ensemble process will overly smooth out the modes by probability averaging. 

By contrast, EBBS allows each ensemble component to explore its own mode (Lines 4--11, Algorithm~\ref{alg:ebbs}). In Figure~\ref{fig:dynamics}, for example, the top sequence yields two plausible next tokens, suggested by each component in the lower level; their probabilities are not smoothed out in our approach, unlike averaging ensembles. The upper level performs soft voting (Lines 12--19, Algorithm~\ref{alg:ebbs}) so as to maintain tractable inference. 

\subsection{EBBS-Based Distillation} \label{subsec:kd}

{\setlength{\algomargin}{1.5em}
\setlength{\columnsep}{20pt}
\setlength{\intextsep}{0pt}

\begin{algorithm}
    \caption{Our EBBS Algorithm}
    \label{alg:ebbs}
    \KwInput{
    $\mathbf x$: input sentence; $Z$: beam size \\
    $K$: number of scorers; \, $p_1, \cdots, p_K$: scorers
    }
    $H \gets \emptyset$ \Comment{candidate outputs} \\
    $\overline{B}_0 \gets \{\langle \textsc{bos},  1 \rangle\}$ \Comment{upper-level beam} \\
    \For{$t = 1, 2, \cdots$}{
        $\triangleright$ lower: individual beam search \\
        \For{$\langle \mathbf y_{1:t-1}, p \rangle \in \overline{B}_{t-1}$}{
            \For{$k = 1,\cdots,K$}{
                $\underline{B}_{t,k} \gets \emptyset$ \Comment{ lower-level beam} \\
                \For{$\mathrm y \in V$}{
                    $p' \gets p_k(\mathrm y|\mathbf y_{1:t-1}, \mathbf x)$ \\
                    $\underline{B}_{t,k}.\operatorname{add}(\langle\mathbf y_{1:t-1} \oplus \mathrm y, p\cdot p'\rangle)$
                }
                $\underline{B}_{t,k} \gets \underline{B}_{t,k}.\operatorname{top}(Z)$
            }
        }
        $\triangleright$ upper: beam synchronization \\
        $D \gets \text{empty dictionary}$ \\
        \For{$k =1,\cdots,K$}{
            \For{$\langle \mathbf y, p\rangle \in \underline{B}_{t, k}$}{
                \eIf{$\mathbf y \in D$}{
                    $D[\mathbf y] \gets p+ D[\mathbf y]$
                }{
                    $D[\mathbf y] \gets p$
                }
            }
        }
        $\overline{B}_t \gets D.\operatorname{top}(Z)$ \\
        $\triangleright$ check for termination \\
        \For{$\langle \mathbf y, p \rangle \in \underline{B}_t$}{
            \If{$\mathrm y_{t} = \textsc{eos}$}  {
                $H.\operatorname{add}(\langle \mathbf y, p \rangle)$ \\
                \If{$|H| = Z$}{
                    \textbf{return} $H.\operatorname{top}(1)$
                }
            }
        }
    }
\end{algorithm}

To improve inference efficiency, we perform knowledge distillation based on the outputs of our EBBS algorithm. 
In particular, we follow~\cite{kim-rush-2016-sequence} and apply a sequence-level knowledge distillation loss, treating the output  $\hat{\mathbf y}$ of our ensemble (serving as a \textit{teacher}) as the pseudo-groundtruth for finetuning the multilingual translation model (serving as a \textit{student}):
\begin{align}
    \mathcal L_{\text{KD}} = - \sum_{t=1}^{|\hat{\mathbf y}|} \log p(\hat {\mathrm y}_t|\hat{\mathbf y}_{1:t-1}, \mathbf x)
\end{align}

Our distilling method is an ensemble-then-distill process. This differs from a straightforward practice of multi-teacher distillation, where the student learns from the union of teachers' outputs~\cite{wu-etal-2021-one}. The commonly applied cross-entropy loss is known to yield overly smooth distributions~\cite{wen2022equal,wen-etal-2023-f}, and the problem becomes more severe with multiple teachers, leading to less satisfactory performance of union distillation~\cite{shayegh2023ensemble}. On the contrary, our approach provides the student with a consolidated pseudo-groundtruth translation, causing less confusion during the distillation process especially when teachers disagree. 

\section{Experiments}

\subsection{Settings}

We evaluated EBBS on two popular benchmark datasets for zero-shot machine translation: IWSLT~\citep{cettolo-etal-2017-overview}, which contains~4 languages (with English) and 6 zero-shot directions; and \mbox{Europarl v7~\citep{koehn-2005-europarl}}, which contains 9 languages and 56 zero-shot directions.

We used BLEU scores~\cite{papineni-etal-2002-bleu} (in particular, SacreBLEU~\cite{post-2018-call}) as our main evaluation metric,\footnote{
We use BLEU$n$ to denote the $n$-gram overlap and BLEU to denote the brevity-penalized geometric mean of BLEU$n$ for $n=1,\cdots,4$. The exact evaluation scripts are available in our codebase (Footnote 1).
} which is one of the most widely used metrics for translation~\cite{JMLR:v22:20-1307,workshop2022bloom}.
For in-depth analyses, we further adopted other popular translation metrics, including the character-level $n$-gram F score (chrF2++)~\cite{popovic-2017-chrf}, the translation edit rate (TER)~\cite{snover-etal-2006-ter}, and a more recent, neural network-based metric called COMET~\cite{rei-etal-2020-comet}.

We replicated \cite{liu-etal-2021-improving-zero} and trained a multilingual translation system as our base model.
Specifically, the neural architecture in \cite{liu-etal-2021-improving-zero} is a 5-layer encoder--decoder Transformer for IWSLT, but has 8 layers for Europarl to accommodate more training data and languages.

For EBBS, we used a beam size of five for both upper- and lower-level beams. In our experiment, we implemented standard beam search for comparison, where we also used a beam size of five, following the common practice~\cite{meister-etal-2020-beam}. A comprehensive beam analysis can be found in our appendix.

\subsection{Competing Methods}

We comprehensively compare our EBBS with direct/pivot translation and other ensemble methods.

\textbf{Direct/pivot translation.} For direct translation, we applied beam search on the multilingual model to translate in unseen directions.
For pivot translation~\citep{wu-wang-2007-pivot,wu-wang-2009-revisiting,vamvas-sennrich-2022-nmtscore}, we used English as the pivot because we have parallel data for translations both from and to English.

\textbf{{Word-level averaging ensemble.}} Averaging is one of the most widely used ensemble techniques in text generation~\citep{sennrich-etal-2016-edinburgh,Freitag2017EnsembleDF,vishnu-kudlu-shanbhogue-etal-2023-improving}. Essentially, the ensemble components' probabilities are first averaged before being fed to the standard beam search.

\textbf{{Word-level voting ensemble.}} The voting ensemble, common in classification tasks, picks the output class based on the number of votes from ensemble components (given by $\operatorname{argmax}$).
However, voting is not common in text generation, because $\operatorname{argmax}$ may select completely different words by the ensemble components due to the large vocabulary size, making voting ineffective.
As a remedy, we pick the word by the highest probability when there is a tie for votes.

\textbf{{Sequence-level voting ensemble.}} Minimum Bayes risk (MBR) decoding is originally designed as a single-model decoding algorithm, where it selects a sequence from a set of beam search results based on similarity~\citep{eikema-aziz-2020-map,muller-sennrich-2021-understanding}. Here, we use it as a sequence-level ensemble technique, where the candidates are the output sequences from different ensemble components.
Let $C = \{\mathbf y_1, \cdots, \mathbf y_K\}$ be the set of candidate outputs given by $K$ ensemble components. The best output is selected as
\begin{align}
    \mathbf y^* = \operatorname*{argmax}\limits_{\mathbf y \in C} \sum_{\mathbf y' \in C \setminus \{\mathbf y\} } \text{BLEU}(\mathbf y, \mathbf y')
\end{align}
where $\text{BLEU}(\mathbf h, \mathbf r)$ computes the BLEU score between a hypothesis $\mathbf h$ and a reference $\mathbf r$.
In essence, MBR selects an output that resembles others most, using BLEU as the similarity metric.

\subsection{Results and Analysis} \label{subsec:results}

\begin{table*}[t]
\centering
\resizebox{\textwidth}{!}{
\begin{tabular}{|c|c|l|rrrrrrrr}
\cline{1-10}
\multirow{9}{*}{IWSLT} & \# & Method & \multicolumn{1}{c}{Average} & \multicolumn{1}{c}{it-nl} & \multicolumn{1}{c}{it-ro} & \multicolumn{1}{c}{nl-it} & \multicolumn{1}{c}{nl-ro} & \multicolumn{1}{c}{ro-it} & \multicolumn{1}{c|}{ro-nl} &  \\ \cline{2-10}
 & 1 & Direct translation~\citep{liu-etal-2021-improving-zero}$^\dagger$ & 17.7 & 18.5 & 17.8 & 17.9 & 15.5 & 19.6 & \multicolumn{1}{r|}{16.8} &  \\
 & 2 & Direct translation (our replication) & 17.29 & 17.46 & \ul{17.48} & 18.23 & \ul{14.63} & 19.65 & \multicolumn{1}{r|}{16.26} &  \\ 
 & 3 & Pivoting (en) & 16.19 & 17.49 & 15.09 & 16.79 & 13.05 & 18.34 & \multicolumn{1}{r|}{16.37} &  \\ \cline{3-10}
 & 4 & Word-level averaging ensemble & 17.28 & 17.29 & 17.44 & 18.33 & 14.65 & 19.69 & \multicolumn{1}{r|}{16.30} &  \\
 & 5 & Word-level voting ensemble & 16.99 & 17.58 & 16.38 & 17.78 & 14.13 & 19.21 & \multicolumn{1}{r|}{16.84} &  \\
 & 6 & Sequence-level voting ensemble (MBR) & 16.72 & 16.64 & 16.53 & 17.83 & 13.74 & 19.48 & \multicolumn{1}{r|}{16.08} &  \\
 & 7 & EBBS (ours) & \ul{18.24} & \ul{19.52} & 17.09 & \ul{19.06} & 14.58 & \ul{20.75} & \multicolumn{1}{r|}{\ul{18.45}} &  \\
 & 8 & Direct w/ EBBS distillation (ours) & \textbf{18.92} & \textbf{19.86} & \textbf{18.80} & \textbf{19.73} & \textbf{15.39} & \textbf{21.23} & \multicolumn{1}{r|}{\textbf{18.48}} & \multicolumn{1}{l}{} \\ \cline{1-10} \noalign{\vskip 3pt} \cline{1-11}
\multirow{9}{*}{Europarl} & \# & Method & \multicolumn{1}{c}{Average} & \multicolumn{1}{c}{da-de} & \multicolumn{1}{c}{da-es} & \multicolumn{1}{c}{da-fi} & \multicolumn{1}{c}{da-fr} & \multicolumn{1}{c}{da-it} & \multicolumn{1}{c}{da-nl} & \multicolumn{1}{c|}{da-pt} \\ \cline{2-11} 
 & 1 & Direct translation~\citep{liu-etal-2021-improving-zero}$^\dagger$ & 26.9 & 24.2 & 33.1 & 18.1 & 30.6 & 26.1 & 26.3 & \multicolumn{1}{r|}{29.9} \\
 & 2 & Direct translation (our replication) & 27.74 & 26.24 & 33.64 & 18.95 & 31.01 & 26.58 & 27.36 & \multicolumn{1}{r|}{30.38} \\ 
 & 3 & Pivoting (en) & 27.69 & 25.17 & 33.87 & 18.70 & 31.44 & 27.12 & 26.75 & \multicolumn{1}{r|}{30.79} \\ \cline{3-11} 
 & 4 & Word-level averaging ensemble & 27.76 & 26.13 & 33.72 & 18.91 & 31.01 & 26.67 & 27.39 & \multicolumn{1}{r|}{30.50} \\
 & 5 & Word-level voting ensemble & 27.45 & 25.76 & 33.24 & 18.39 & 30.96 & 26.83 & 26.63 & \multicolumn{1}{r|}{30.37} \\
 & 6 & Sequence-level voting ensemble (MBR) & 27.90 & 25.90 & 33.95 & 19.15 & 31.50 & 27.15 & 27.09 & \multicolumn{1}{r|}{30.55} \\
 & 7 & EBBS (ours) & \ul{28.36} & \ul{26.32} & \ul{34.28} & \ul{19.43} & \ul{31.97} & \ul{27.67} & \textbf{27.78} & \multicolumn{1}{r|}{\ul{31.08}} \\
 & 8 & Direct w/ EBBS distillation (ours) & \textbf{28.54} & \textbf{26.75} & \textbf{34.68} & \textbf{19.89} & \textbf{32.00} & \textbf{27.69} & \ul{27.61} & \multicolumn{1}{r|}{\textbf{31.19}} \\ \hline
\end{tabular}
}
\caption{Main results of BLEU scores on IWSLT and Europarl.
The best results are in \textbf{bold}; the second best results are \ul{underlined}. $^\dagger$ indicates cited results; others were obtained by our experimentation. 
}
\label{tab:main}
\end{table*}

\textbf{Main results.}
Our experiment starts by a replication of the base multilingual model \cite{liu-etal-2021-improving-zero}. As shown in \mbox{Rows~1--2}, Table~\ref{tab:main}, the results are generally close, indicating that our replication is successful and ready for ensemble research. Further, we tried English pivoting (Row 3), a common zero-shot translation method.
In our experiments, we find that it does not outperform direct translation, as pivoting methods may suffer from the error accumulation problem due to two-step translation.

We then compare different ensemble techniques, including our proposed EBBS. 
We notice that IWSLT contains four languages (with English); thus we have two available pivoting directions (excluding source and target), which, along with direct translation, are our three ensemble components. For Europarl, it contains nine languages; for performance and efficiency concerns (to be shown in Figure~\ref{fig:scaling}), we also consider three translation paths as our ensemble components: direction translation, English pivoting, and a second pivot.\footnote{We use the first available language in the order of Spanish, German, and French. For example, Spanish-to-German translation will have to use French as the pivot. These languages are chosen because they have the most content on the Internet according to the Web Technology Surveys (\url{https://w3techs.com/technologies/overview/content_language}).}

We study the common ensemble technique of word-level averaging (Row~4), which has been used in previous translation research~\citep{Freitag2017EnsembleDF}.
As we can see, the averaging ensemble performs similarly to direct translation on both datasets.
Our zero-shot results are different from~\cite{Freitag2017EnsembleDF}, which shows a word-level averaging ensemble of random seeds can improve performance in the supervised setting. This is because models trained with different random seeds exhibit similar behavior, and averaging their probabilities achieves a denoising effect. However, our ensemble components differ drastically in terms of their strengths and expertise due to the different translation paths (direct and pivot translations).
Thus, word averaging fails to improve translation quality in our setting.

Alternatively, voting ensembles can also be applied, at either the word level or the sequence level.
As seen, word-level voting is not effective, as it is worse than direct translation on both datasets (Row~5).
This is expected because the voted words (top predictions) by the ensemble components may not overlap due to the large vocabulary size.
In such cases, the algorithm defaults to choosing the word with the highest probability, causing the ensemble to follow the most peaked distributions.

Sequence-level voting should also be done in a soft manner, and minimum Bayes risk (MBR) decoding can be thought of as using a Bayes risk to softly ``vote'' the candidate outputs. As seen from Row 6, such a method works relatively well on Europarl, achieving the second-highest performance across all ensemble methods; however, it works poorly on the IWSLT dataset.
The main drawback of sequence-level voting is that it can only \textit{select} one of the ensemble components' output.
This may not work well when the individual ensemble components are weak, especially with the small IWSLT dataset.
Such a selective sequence-level ensemble cannot integrate different expertise of its components during generation.

\setlength{\columnsep}{10pt}
\begin{table}
\centering
\resizebox{0.9\columnwidth}{!}{
\begin{tabular}{|l|lccc|}
\hline
Dataset & \multicolumn{1}{l|}{Method} & Avg. BLEU & Wins & Losses \\ \hline \hline
\multirow{2}{*}{IWSLT} & \multicolumn{1}{l|}{Direct translation} & 17.29 & 2 & 4 \\
 & \multicolumn{1}{l|}{EBBS (ours)} & \textbf{18.24} & \textbf{4} & \textbf{2} \\ \hline
\multirow{2}{*}{Europarl} & \multicolumn{1}{l|}{Direct translation} & 27.85 & 4 & 52 \\
 & \multicolumn{1}{l|}{EBBS (ours)} & \textbf{28.44} & \textbf{52} & \textbf{4} \\ \hline\hline
\multirow{2}{*}{Overall} & \multicolumn{1}{l|}{Direct translation} & 26.83 & 6 & 56 \\
 & \multicolumn{1}{l|}{EBBS (ours)} & \textbf{27.45} & \textbf{56} & \textbf{6} \\ \hline
$p$-value & \multicolumn{4}{c|}{3e-11} \\ \hline
\end{tabular}
}
\caption{Pairwise comparison on all 62 zero-shot directions in both datasets. The $p$-value is given by a two-sided binomial test.}
\label{tab:europarl-summary}
\end{table}

Unlike existing ensemble methods, our EBBS algorithm achieves higher performance in most directions on both datasets. Noticing that Europarl contains 56 zero-shot directions, we could only present in Table~\ref{tab:main} the first seven directions based on the order provided by the dataset, due to the space limit. Table~\ref{tab:europarl-summary} further shows a pairwise comparison against direct translation (a strong baseline in our experiment) in all zero-shot directions. 
As seen, EBBS achieves higher performance in 56 out of 62 cases across two datasets, showing strong statistical evidence for its effectiveness, with a $p$-value of 3e-11 in a two-sided binomial test.

We also evaluate EBBS-based distillation (Row~8, Table~\ref{tab:main}).
Again, since Europarl has 56 zero-shot directions, we follow the standard practice~\cite{JMLR:v22:20-1307} and select a subset of directions, namely, Danish to other languages, to save computational cost.
As seen in Row~8, EBBS-based distillation consistently achieves the highest performance in all directions (except for Danish-to-Dutch translation).
This shows that an EBBS-distilled model can outperform EBBS, which is not surprising because learning can smooth out the noise of various heuristics~\cite{deshmukh-etal-2021-unsupervised-chunking,Jolly_Zhang_Dengel_Mou_2022}, such as the ensemble algorithm in our scenario.
Importantly, EBBS-based distillation achieves significantly higher translation quality with \textit{no inference overhead} compared with direct translation.

\textbf{Distillation analysis.}
We compare EBBS-based distillation with other distilling methods.
Here, we only focus on Italian-to-Dutch\footnote{We could only afford one translation direction for this analysis, because we need to train different models for all competing distilling methods. This differs from Table~\ref{tab:main}, where we follow previous work and perform EBBS-based distillation for Danish to other languages. We chose Italian-to-Dutch translation here, because it is the first one in IWSLT, conveniently also available in Europarl.} translation to save computational cost.

In particular, we consider two alternative distilling methods: direct and union distillation. Direct distillation finetunes the multilingual model with its own predictions based on direct translation.
Union distillation, on the other hand, takes the union of the teachers' outputs (direct and pivot translations) for training, which is under a controlled experimental setup, because it uses exactly the same translation paths as our EBBS-based distillation.

As seen in Table~\ref{tab:distill}, both direct and union distillation marginally improve the performance compared with no distillation. Intriguingly, learning from the union of multiple teachers is not necessarily better than learning from the best teacher (namely, direct translation). This is because multiple teachers may provide conflicting training signals and confuse the student model. 

On the contrary, our EBBS-based distillation consistently outperforms direct and union distillation on both datasets. This shows that our ensemble-then-distill approach is able to consolidate the knowledge of multiple teachers to better train the student model.

\begin{table*}[!t]
\centering
\resizebox{\linewidth}{!}{
\begin{tabular}{|l|ll|cccccccc|}
\hline
Dataset & \multicolumn{2}{c|}{Method} & BLEU$^\uparrow$ & BLEU1$^\uparrow$ & BLEU2$^\uparrow$ & BLEU3$^\uparrow$ & BLEU4$^\uparrow$ & chrF2++$^\uparrow$ & TER$^\downarrow$ & COMET$^\uparrow$ \\ \hline
\multirow{5}{*}{IWSLT} & \multicolumn{2}{l|}{EBBS} & 19.52 & 51.87 & 25.12 & 13.88 & 8.02 & 45.63 & 71.36 & 0.7341 \\ \cline{2-11} 
 & \multicolumn{1}{l|}{\multirow{4}{*}{\begin{tabular}[c]{@{}l@{}}Direct \\ Translation\end{tabular}}} & No distillation & 17.46 & \ul{50.49} & 23.01 & 12.01 & 6.66 & 43.73 & \ul{72.02} & 0.7088 \\
 & \multicolumn{1}{l|}{} & Direct distillation & \ul{18.10} & 50.37 & \ul{23.53} & \ul{12.63} & \ul{7.17} & 44.48 & 72.86 & 0.7144 \\
 & \multicolumn{1}{l|}{} & Union distillation & 17.80 & 49.21 & 23.01 & 12.51 & 7.10 & \ul{44.93} & 75.92 & \ul{0.7221} \\
 & \multicolumn{1}{l|}{} & EBBS distillation & \textbf{20.13} & \textbf{53.20} & \textbf{26.06} & \textbf{14.33} & \textbf{8.26} & \textbf{46.46} & \textbf{69.28} & \textbf{0.7428} \\ \hline \hline
\multirow{5}{*}{Europarl} & \multicolumn{2}{l|}{EBBS} & 26.10 & 57.07 & 31.00 & 19.76 & 13.28 & 52.75 & 65.63 & 0.8340 \\ \cline{2-11} 
 & \multicolumn{1}{l|}{\multirow{4}{*}{\begin{tabular}[c]{@{}l@{}}Direct \\ Translation\end{tabular}}} & No distillation & 25.33 & 56.32 & 30.08 & 19.01 & 12.78 & 52.32 & 66.56 & 0.8276 \\
 & \multicolumn{1}{l|}{} & Direct distillation & 25.44 & 56.54 & 30.28 & 19.13 & 12.79 & 52.61 & 66.34 & \ul{0.8286} \\
 & \multicolumn{1}{l|}{} & Union distillation & \ul{25.53} & \ul{56.58} & \ul{30.34} & \ul{19.18} & \ul{12.91} & \ul{52.63} & \ul{66.27} & 0.8282 \\
 & \multicolumn{1}{l|}{} & EBBS distillation & \textbf{25.92} & \textbf{56.76} & \textbf{30.68} & \textbf{19.57} & \textbf{13.24} & \textbf{52.73} & \textbf{66.04} & \textbf{0.8307} \\ \hline
\end{tabular}
}
\caption{Comparison of various distilling methods for Italian-to-Dutch translation. $^{\uparrow/\downarrow}$The higher/lower, the better.}
\label{tab:distill}
\end{table*}

\begin{table}[]
\centering
\resizebox{0.9\columnwidth}{!}{
\begin{tabular}{|l|rrrr|}
\hline
Method & \multicolumn{1}{c}{BLEU$^\uparrow$} & \multicolumn{1}{c}{chrF2++$^\uparrow$} & \multicolumn{1}{c}{TER$^\downarrow$} & \multicolumn{1}{c|}{COMET$^\uparrow$} \\ \hline
Direct translation & \ul{25.33} & \ul{52.32} & 66.56 & 0.8276 \\
Pivoting (en) & 25.08 & 51.92 & \ul{66.24} & \ul{0.8322} \\
Pivoting (es) & 24.40 & 51.71 & 67.91 & 0.8192 \\
Pivoting (pt) & 24.34 & 51.61 & 67.68 & 0.8191 \\
Pivoting (fr) & 24.20 & 51.61 & 67.84 & 0.8208 \\
Pivoting (de) & 23.65 & 50.70 & 67.89 & 0.8157 \\
Pivoting (da) & 23.12 & 50.36 & 69.00 & 0.8156 \\
Pivoting (fi) & 20.74 & 48.11 & 70.59 & 0.8051 \\
Our EBBS & \textbf{26.10} & \textbf{52.75} & \textbf{65.63} & \textbf{0.8340} \\ \hline
\end{tabular}
}
\caption{The performance of direct/pivot translation and our EBBS for Italian-to-Dutch translation on Europarl.}
\label{tab:components}
\end{table}

Further, the analysis suggests that our EBBS-distilled model achieves a speedup of multiple times compared with EBBS, because after distillation the model is used by direct translation. This is a significant result, because our EBBS-based distillation not only speeds up the EBBS ensemble approach, but also improves the translation quality of EBBS as shown in Row 8, Table~\ref{tab:main}.

\textbf{Analysis of ensemble components.} In Table~\ref{tab:components}, we analyze the ensemble components to better understand our ensemble technique for zero-shot machine translation. As seen, direct translation is an effective approach, which is consistent with previous literature~\cite{JMLR:v22:20-1307,liu-etal-2021-improving-zero}. English pivoting achieves higher performance for some metrics but lower for others; it is not conclusively better than direct translation, probably because of the error accumulation problem. Pivoting through non-English languages degrades the performance to a large extent because lacking supervision along the pivoting path leads to two steps of zero-shot translation.
EBBS, on the other hand, combines the strengths of individual components and consistently outperforms them in all metrics.

\begin{figure}[t]
    \centering
    \includegraphics[width=0.8\columnwidth]{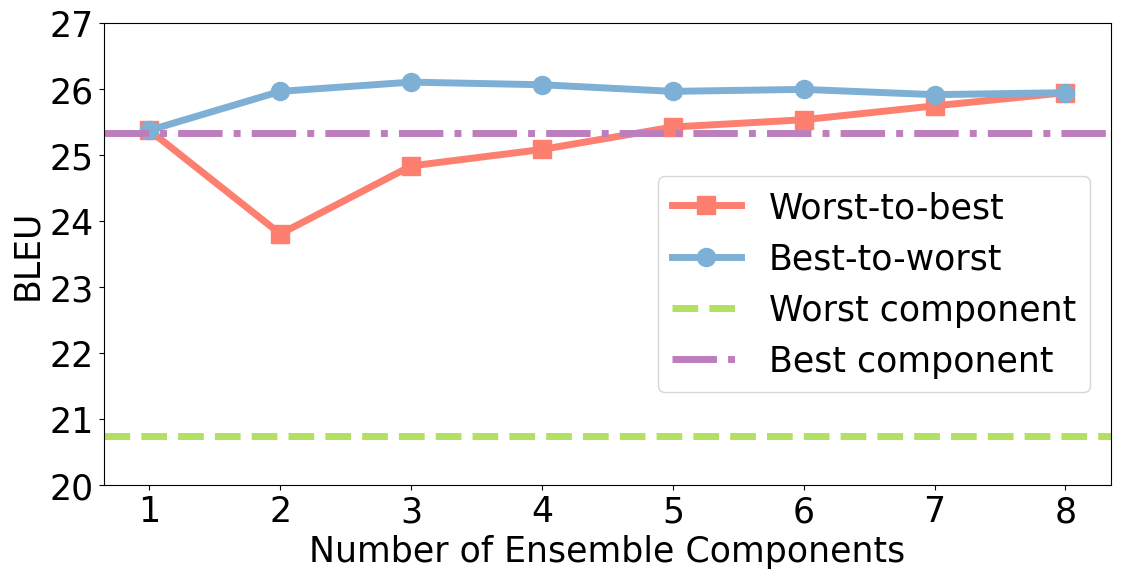}
    \caption{Analysis of the number of ensemble components for Italian-to-Dutch translation on Europarl.}
    \label{fig:scaling}
\end{figure}

We further study how EBBS performs with different numbers of ensemble components.
Specifically, we analyze two incremental ensemble settings: best-to-worst and worst-to-best. In both cases, we start with direct translation; then we incrementally add the next ``best'' or ``worst'' pivot translation according to Table~\ref{tab:components}.

Figure~\ref{fig:scaling} shows the trends of incremental ensembles. If we add the best pivot directions, the performance peaks at three ensemble components; interestingly, the inclusion of weaker components does not affect EBBS much.
On the other hand, adding the worst pivot translation at the beginning leads to an immediate drop of 1.6 BLEU points, which then largely recovers with the second pivot.
This is reasonable because the worst pivot (Finnish) is 4.6 BLEU points lower than direct translation, and EBBS cannot decide on which of the two ensemble components to trust; despite this, the performance of EBBS is still much better than the average performance of the components.
With a second pivot, there is a third ``opinion'' when the first two components ``disagree.''
The performance continues to rise if more and stronger components are added. 
In fact, our ensemble even surpasses the baseline with 4 weakest pivot translations, each of which is at least 1 BLEU point lower than the baseline. This demonstrates that EBBS is flexible and works well with both strong and weak ensemble components.

\textbf{Appendix.} We present additional details and results in the appendix:
\begin{compactitem}
\item[\quad A.] Beam search,
\item[\quad B.] Experimental details,
\item[\quad C.] Analysis of inference efficiency,
\item[\quad D.] Average performance across tasks,
\item[\quad E.] Analysis of beam size,
\item[\quad F.] Entropy of distilled models,
\item[\quad G.] Analysis of voting methods in EBBS, and
\item[\quad H.] Case study.
\end{compactitem}

\section{Conclusion}
In this work, we address ensemble-based zero-shot machine translation by directly translating and pivoting through different languages. We further design a novel bi-level beam search algorithm (called EBBS) for decoding. We evaluated EBBS on two popular zero-shot translation datasets, IWSLT and Europarl. 
Results show that EBBS outperforms existing ensemble techniques, and that the high-quality translations produced by EBBS can be used for distillation to improve translation efficiency (and sometimes also output quality).

\FloatBarrier

\section*{Acknowledgments}
The research is supported in part by the Natural Sciences and Engineering Research Council of Canada (NSERC), a Mitacs Accelerate project, the Amii Fellow Program, the Canada CIFAR AI Chair Program, an Alberta Innovates Program, and the Digital Research Alliance of Canada (alliancecan.ca).
We used the open-domain artwork ``The Great Wave off Kanagawa’' by Katsushika Hokusai in the title and as the background for Figure~\ref{fig:dynamics}.

\FloatBarrier

\bibliography{aaai25}

\begin{thebibliography}{77}
\providecommand{\natexlab}[1]{#1}

\bibitem[{Babych, Hartley, and Sharoff(2007)}]{babych-etal-2007-translating}
Babych, B.; Hartley, A.; and Sharoff, S. 2007.
\newblock Translating from under-resourced languages: Comparing direct transfer against pivot translation.
\newblock In \emph{MTSummit}.

\bibitem[{Bahdanau, Cho, and Bengio(2015)}]{bahdanau2014neural}
Bahdanau, D.; Cho, K.; and Bengio, Y. 2015.
\newblock Neural machine translation by jointly learning to align and translate.
\newblock In \emph{ICLR}.

\bibitem[{Baziotis et~al.(2022)Baziotis, Artetxe, Cross, and Bhosale}]{baziotis-etal-2022-multilingual}
Baziotis, C.; Artetxe, M.; Cross, J.; and Bhosale, S. 2022.
\newblock Multilingual machine translation with hyper-adapters.
\newblock In \emph{EMNLP}, 1170--1185.

\bibitem[{Bickel and Doksum(2015)}]{MBRbook}
Bickel, P.~J.; and Doksum, K.~A. 2015.
\newblock \emph{Mathematical Statistics: Basic Ideas and Selected Topics}.
\newblock CRC Press.

\bibitem[{Breiman(1996)}]{breiman1996bagging}
Breiman, L. 1996.
\newblock Bagging predictors.
\newblock \emph{Machine Learning}, 24: 123--140.

\bibitem[{Brown et~al.(1990)}]{brown-etal-1990-statistical}
Brown, P.~F.; et~al. 1990.
\newblock A statistical approach to machine translation.
\newblock \emph{CL}, 16(2): 79--85.

\bibitem[{B{\"u}hlmann and Yu(2002)}]{buhlmann2002analyzing}
B{\"u}hlmann, P.; and Yu, B. 2002.
\newblock Analyzing bagging.
\newblock \emph{The Annals of Statistics}, 30(4): 927--961.

\bibitem[{Cettolo et~al.(2017)}]{cettolo-etal-2017-overview}
Cettolo, M.; et~al. 2017.
\newblock Overview of the {IWSLT} 2017 evaluation campaign.
\newblock In \emph{IWSLT}, 2–14.

\bibitem[{Chronopoulou, Stojanovski, and Fraser(2023)}]{chronopoulou-etal-2023-language}
Chronopoulou, A.; Stojanovski, D.; and Fraser, A. 2023.
\newblock Language-family adapters for low-resource multilingual neural machine translation.
\newblock In \emph{Proceedings of the Workshop on Technologies for Machine Translation of Low-Resource Languages}, 59--72.

\bibitem[{Dabre, Chu, and Kunchukuttan(2020)}]{10.1145/3406095}
Dabre, R.; Chu, C.; and Kunchukuttan, A. 2020.
\newblock A survey of multilingual neural machine translation.
\newblock \emph{ACM Computing Surveys}, 53(5).

\bibitem[{Dabre and Kurohashi(2017)}]{DBLP:journals/corr/abs-1710-01025}
Dabre, R.; and Kurohashi, S. 2017.
\newblock {MMCR4NLP}: Multilingual multiway corpora repository for natural language Processing.
\newblock \emph{arXiv preprint arXiv:1710.01025}.

\bibitem[{Deshmukh et~al.(2021)Deshmukh, Zhang, Li, Lin, and Mou}]{deshmukh-etal-2021-unsupervised-chunking}
Deshmukh, A.~A.; Zhang, Q.; Li, M.; Lin, J.; and Mou, L. 2021.
\newblock Unsupervised chunking as syntactic structure induction with a knowledge-transfer approach.
\newblock In \emph{EMNLP Findings}, 3626--3634.

\bibitem[{Dong et~al.(2020)Dong, Yu, Cao, Shi, and Ma}]{dong2020survey}
Dong, X.; Yu, Z.; Cao, W.; Shi, Y.; and Ma, Q. 2020.
\newblock A survey on ensemble learning.
\newblock \emph{Frontiers of Computer Science}, 14: 241--258.

\bibitem[{Dugast, Senellart, and Koehn(2007)}]{dugast-etal-2007-statistical}
Dugast, L.; Senellart, J.; and Koehn, P. 2007.
\newblock Statistical post-editing on {SYSTRAN}'s rule-based translation system.
\newblock In \emph{WMT}, 220--223.

\bibitem[{Eikema and Aziz(2020)}]{eikema-aziz-2020-map}
Eikema, B.; and Aziz, W. 2020.
\newblock Is {MAP} decoding all you need? {The} inadequacy of the mode in neural machine translation.
\newblock In \emph{COLING}, 4506--4520.

\bibitem[{Fan et~al.(2021)}]{JMLR:v22:20-1307}
Fan, A.; et~al. 2021.
\newblock Beyond {E}nglish-centric multilingual machine translation.
\newblock \emph{JMLR}, 22(107): 1--48.

\bibitem[{Forcada et~al.(2011)}]{forcada2011apertium}
Forcada, M.~L.; et~al. 2011.
\newblock Apertium: A free/open-source platform for rule-based machine translation.
\newblock \emph{Machine Translation}, 25: 127--144.

\bibitem[{Freitag, Al-Onaizan, and Sankaran(2017)}]{Freitag2017EnsembleDF}
Freitag, M.; Al-Onaizan, Y.; and Sankaran, B. 2017.
\newblock Ensemble distillation for neural machine translation.
\newblock \emph{arXiv preprint arXiv:1702.01802}.

\bibitem[{Gaikwad et~al.(2024)Gaikwad, Doshi, Dabre, and Bhattacharyya}]{gaikwad2024effective}
Gaikwad, P.; Doshi, M.; Dabre, R.; and Bhattacharyya, P. 2024.
\newblock How effective is multi-source pivoting for translation of low resource {I}ndian languages?
\newblock \emph{arXiv preprint arXiv:2406.13332}.

\bibitem[{Ganaie et~al.(2022)Ganaie, Hu, Malik, Tanveer, and Suganthan}]{GANAIE2022105151}
Ganaie, M.; Hu, M.; Malik, A.; Tanveer, M.; and Suganthan, P. 2022.
\newblock Ensemble deep learning: A review.
\newblock \emph{Engineering Applications of Artificial Intelligence}, 115: 105151.

\bibitem[{Gu et~al.(2019)Gu, Wang, Cho, and Li}]{gu-etal-2019-improved}
Gu, J.; Wang, Y.; Cho, K.; and Li, V.~O. 2019.
\newblock Improved zero-shot neural machine translation via ignoring spurious Correlations.
\newblock In \emph{ACL}, 1258--1268.

\bibitem[{Hasselt(2010)}]{NIPS2010_091d584f}
Hasselt, H. 2010.
\newblock Double {Q}-learning.
\newblock In \emph{NeurIPS}.

\bibitem[{Hastie et~al.(2009)Hastie, Rosset, Zhu, and Zou}]{hastie2009multi}
Hastie, T.; Rosset, S.; Zhu, J.; and Zou, H. 2009.
\newblock Multi-class {A}da{B}oost.
\newblock \emph{Statistics and Its Interface}, 2(3): 349--360.

\bibitem[{He et~al.(2023)}]{he-etal-2023-gradient}
He, D.; et~al. 2023.
\newblock Gradient-based gradual pruning for language-specific multilingual neural machine translation.
\newblock In \emph{EMNLP}, 654--670.

\bibitem[{Hendy et~al.(2021)}]{hendy-etal-2021-ensembling}
Hendy, A.; et~al. 2021.
\newblock Ensembling of distilled models from multi-task teachers for constrained resource language pairs.
\newblock In \emph{WMT}, 130--135.

\bibitem[{Holtzman et~al.(2019)Holtzman, Buys, Du, Forbes, and Choi}]{holtzman2019curious}
Holtzman, A.; Buys, J.; Du, L.; Forbes, M.; and Choi, Y. 2019.
\newblock The curious case of neural text degeneration.
\newblock In \emph{ICLR}.

\bibitem[{Johnson et~al.(2017)}]{johnson-etal-2017-googles}
Johnson, M.; et~al. 2017.
\newblock {G}oogle{'}s multilingual neural machine translation system: Enabling zero-shot translation.
\newblock \emph{TACL}, 5: 339--351.

\bibitem[{Jolly et~al.(2022)Jolly, Zhang, Dengel, and Mou}]{Jolly_Zhang_Dengel_Mou_2022}
Jolly, S.; Zhang, Z.~X.; Dengel, A.; and Mou, L. 2022.
\newblock Search and learn: Improving semantic coverage for data-to-text generation.
\newblock In \emph{AAAI}, 10858--10866.

\bibitem[{Kim and Rush(2016)}]{kim-rush-2016-sequence}
Kim, Y.; and Rush, A.~M. 2016.
\newblock Sequence-level knowledge distillation.
\newblock In \emph{EMNLP}, 1317--1327.

\bibitem[{Kobayashi(2018)}]{kobayashi-2018-frustratingly}
Kobayashi, H. 2018.
\newblock Frustratingly easy model ensemble for abstractive summarization.
\newblock In \emph{EMNLP}, 4165--4176.

\bibitem[{Koehn(2005)}]{koehn-2005-europarl}
Koehn, P. 2005.
\newblock {E}uroparl: A parallel corpus for statistical machine translation.
\newblock In \emph{MTSummit}, 79--86.

\bibitem[{Koehn(2009)}]{koehn2009statistical}
Koehn, P. 2009.
\newblock \emph{Statistical Machine Translation}.
\newblock Cambridge University Press.

\bibitem[{Lample et~al.(2018{\natexlab{a}})Lample, Conneau, Denoyer, and Ranzato}]{lample2018unsupervised}
Lample, G.; Conneau, A.; Denoyer, L.; and Ranzato, M. 2018{\natexlab{a}}.
\newblock Unsupervised machine translation using monolingual corpora only.
\newblock In \emph{ICLR}.

\bibitem[{Lample et~al.(2018{\natexlab{b}})Lample, Ott, Conneau, Denoyer, and Ranzato}]{lample-etal-2018-phrase}
Lample, G.; Ott, M.; Conneau, A.; Denoyer, L.; and Ranzato, M. 2018{\natexlab{b}}.
\newblock Phrase-based {\&} neural unsupervised machine translation.
\newblock In \emph{EMNLP}, 5039--5049.

\bibitem[{Lewis et~al.(2020)}]{lewis-etal-2020-bart}
Lewis, M.; et~al. 2020.
\newblock {BART}: Denoising sequence-to-sequence pre-training for natural language generation, translation, and comprehension.
\newblock In \emph{ACL}, 7871--7880.

\bibitem[{Liu et~al.(2021)Liu, Niehues, Cross, Guzm{\'a}n, and Li}]{liu-etal-2021-improving-zero}
Liu, D.; Niehues, J.; Cross, J.; Guzm{\'a}n, F.; and Li, X. 2021.
\newblock Improving zero-shot translation by disentangling positional information.
\newblock In \emph{ACL-IJCNLP}, 1259--1273.

\bibitem[{Meister, Cotterell, and Vieira(2020)}]{meister-etal-2020-beam}
Meister, C.; Cotterell, R.; and Vieira, T. 2020.
\newblock If beam search is the answer, what was the question?
\newblock In \emph{EMNLP}, 2173--2185.

\bibitem[{Mohammadshahi, Vamvas, and Sennrich(2024)}]{mohammadshahi-etal-2024-investigating}
Mohammadshahi, A.; Vamvas, J.; and Sennrich, R. 2024.
\newblock Investigating multi-pivot ensembling with massively multilingual machine translation models.
\newblock In \emph{Proceedings of the Workshop on Insights from Negative Results in NLP}, 169--180.

\bibitem[{Mohiuddin and Joty(2020)}]{mohiuddin-joty-2020-unsupervised}
Mohiuddin, T.; and Joty, S. 2020.
\newblock Unsupervised word translation with adversarial autoencoder.
\newblock \emph{CL}, 46(2): 257--288.

\bibitem[{Muennighoff et~al.(2023)}]{NEURIPS2023_9d89448b}
Muennighoff, N.; et~al. 2023.
\newblock Scaling data-constrained language models.
\newblock In \emph{NeurIPS}, 50358--50376.

\bibitem[{M{\"u}ller and Sennrich(2021)}]{muller-sennrich-2021-understanding}
M{\"u}ller, M.; and Sennrich, R. 2021.
\newblock Understanding the properties of minimum {B}ayes risk decoding in neural machine translation.
\newblock In \emph{ACL-IJCNLP}, 259--272.

\bibitem[{Natekin and Knoll(2013)}]{natekin2013gradient}
Natekin, A.; and Knoll, A. 2013.
\newblock Gradient boosting machines, a tutorial.
\newblock \emph{Frontiers in Neurorobotics}, 7: 1--21.

\bibitem[{Ott et~al.(2019)}]{ott-etal-2019-fairseq}
Ott, M.; et~al. 2019.
\newblock fairseq: A fast, extensible toolkit for sequence modeling.
\newblock In \emph{NAACL-HLT: Demonstrations}, 48--53.

\bibitem[{Papineni et~al.(2002)Papineni, Roukos, Ward, and Zhu}]{papineni-etal-2002-bleu}
Papineni, K.; Roukos, S.; Ward, T.; and Zhu, W.-J. 2002.
\newblock {BLEU}: A method for automatic evaluation of machine translation.
\newblock In \emph{ACL}, 311--318.

\bibitem[{Popovi{\'c}(2017)}]{popovic-2017-chrf}
Popovi{\'c}, M. 2017.
\newblock chr{F}++: Words helping character n-grams.
\newblock In \emph{WMT}, 612--618.

\bibitem[{Post(2018)}]{post-2018-call}
Post, M. 2018.
\newblock A call for clarity in reporting {BLEU} scores.
\newblock In \emph{WMT}, 186--191.

\bibitem[{Radford et~al.(2019)}]{radford2019language}
Radford, A.; et~al. 2019.
\newblock Language models are unsupervised multitask learners.
\newblock \emph{OpenAI Blog}.

\bibitem[{Raffel et~al.(2020)}]{t52020}
Raffel, C.; et~al. 2020.
\newblock Exploring the limits of transfer learning with a unified text-to-text {T}ransformer.
\newblock \emph{JMLR}, 21(140): 1--67.

\bibitem[{Ranathunga et~al.(2023)Ranathunga, Lee, Prifti~Skenduli, Shekhar, Alam, and Kaur}]{10.1145/3567592}
Ranathunga, S.; Lee, E.-S.~A.; Prifti~Skenduli, M.; Shekhar, R.; Alam, M.; and Kaur, R. 2023.
\newblock Neural machine translation for low-resource languages: A survey.
\newblock \emph{ACM Computing Survey}, 55(11).

\bibitem[{Razmara and Sarkar(2013)}]{razmara-sarkar-2013-ensemble}
Razmara, M.; and Sarkar, A. 2013.
\newblock Ensemble triangulation for statistical machine translation.
\newblock In \emph{IJCNLP}, 252--260.

\bibitem[{Rei et~al.(2020)Rei, Stewart, Farinha, and Lavie}]{rei-etal-2020-comet}
Rei, R.; Stewart, C.; Farinha, A.~C.; and Lavie, A. 2020.
\newblock {COMET}: A neural framework for {MT} evaluation.
\newblock In \emph{EMNLP}, 2685--2702.

\bibitem[{Scao et~al.(2022)}]{workshop2022bloom}
Scao, T.~L.; et~al. 2022.
\newblock B{LOOM}: A 176{B}-parameter open-access multilingual language model.
\newblock \emph{arXiv preprint arXiv:2211.05100}.

\bibitem[{Schapire(2003)}]{schapire2003boosting}
Schapire, R.~E. 2003.
\newblock The boosting approach to machine learning: An overview.
\newblock \emph{Nonlinear Estimation and Classification}, 149--171.

\bibitem[{Sennrich, Haddow, and Birch(2016{\natexlab{a}})}]{sennrich-etal-2016-edinburgh}
Sennrich, R.; Haddow, B.; and Birch, A. 2016{\natexlab{a}}.
\newblock {E}dinburgh neural machine translation systems for {WMT} 16.
\newblock In \emph{WMT}, 371--376.

\bibitem[{Sennrich, Haddow, and Birch(2016{\natexlab{b}})}]{sennrich-etal-2016-neural}
Sennrich, R.; Haddow, B.; and Birch, A. 2016{\natexlab{b}}.
\newblock Neural machine translation of rare words with subword units.
\newblock In \emph{ACL}, 1715--1725.

\bibitem[{Shanbhogue et~al.(2023)Shanbhogue, Xue, Saha, Zhang, and Ganesan}]{vishnu-kudlu-shanbhogue-etal-2023-improving}
Shanbhogue, A. V.~K.; Xue, R.; Saha, S.; Zhang, D.; and Ganesan, A. 2023.
\newblock Improving low resource speech translation with data augmentation and ensemble strategies.
\newblock In \emph{IWSLT}, 241--250.

\bibitem[{Shayegh et~al.(2024)Shayegh, Cao, Zhu, Cheung, and Mou}]{shayegh2023ensemble}
Shayegh, B.; Cao, Y.; Zhu, X.; Cheung, J.~C.; and Mou, L. 2024.
\newblock Ensemble distillation for unsupervised constituency parsing.
\newblock In \emph{ICLR}.

\bibitem[{Shayegh, Wen, and Mou(2024)}]{shayegh-etal-2024-tree}
Shayegh, B.; Wen, Y.; and Mou, L. 2024.
\newblock Tree-averaging algorithms for ensemble-based unsupervised discontinuous constituency parsing.
\newblock In \emph{ACL}, 15135--15156.

\bibitem[{Shayegh et~al.(2025)}]{shayegh-selection}
Shayegh, B.; et~al. 2025.
\newblock Error diversity matters: An error-resistant ensemble method for unsupervised dependency parsing.
\newblock In \emph{AAAI}.

\bibitem[{Snover et~al.(2006)Snover, Dorr, Schwartz, Micciulla, and Makhoul}]{snover-etal-2006-ter}
Snover, M.; Dorr, B.; Schwartz, R.; Micciulla, L.; and Makhoul, J. 2006.
\newblock A study of translation edit rate with targeted human annotation.
\newblock In \emph{Association for Machine Translation in the Americas}, 223--231.

\bibitem[{Vamvas and Sennrich(2022)}]{vamvas-sennrich-2022-nmtscore}
Vamvas, J.; and Sennrich, R. 2022.
\newblock {NMTS}core: A multilingual analysis of translation-based text similarity measures.
\newblock In \emph{EMNLP Findings}, 198--213.

\bibitem[{van Hasselt, Guez, and Silver(2016)}]{Hasselt_Guez_Silver_2016}
van Hasselt, H.; Guez, A.; and Silver, D. 2016.
\newblock Deep reinforcement learning with double {Q}-learning.
\newblock In \emph{AAAI}, 2094--2100.

\bibitem[{Vaswani et~al.(2017)Vaswani, Shazeer, Parmar, Uszkoreit, Jones, Gomez, Kaiser, and Polosukhin}]{vaswani2017-transformer}
Vaswani, A.; Shazeer, N.; Parmar, N.; Uszkoreit, J.; Jones, L.; Gomez, A.~N.; Kaiser, {\L}.; and Polosukhin, I. 2017.
\newblock Attention is all you need.
\newblock In \emph{NeurIPS}.

\bibitem[{Wang and Zhang(2022)}]{Wang_Zhang_2022}
Wang, Q.; and Zhang, J. 2022.
\newblock Parameter differentiation based multilingual neural machine translation.
\newblock In \emph{AAAI}, 11440--11448.

\bibitem[{Wang, Lipton, and Tsvetkov(2020)}]{wang-etal-2020-negative}
Wang, Z.; Lipton, Z.~C.; and Tsvetkov, Y. 2020.
\newblock On negative interference in multilingual models: Findings and a meta-learning treatment.
\newblock In \emph{EMNLP}, 4438--4450.

\bibitem[{Wei et~al.(2019)Wei, Lu, Mou, Zhou, Poupart, Li, and Jin}]{wei2019neural}
Wei, B.; Lu, S.; Mou, L.; Zhou, H.; Poupart, P.; Li, G.; and Jin, Z. 2019.
\newblock Why do neural dialog systems generate short and meaningless replies? {A} comparison between dialog and translation.
\newblock In \emph{ICASSP}, 7290--7294.

\bibitem[{Wen et~al.(2023{\natexlab{a}})Wen, Hao, Cao, and Mou}]{wen2022equal}
Wen, Y.; Hao, Y.; Cao, Y.; and Mou, L. 2023{\natexlab{a}}.
\newblock An equal-size hard {EM} algorithm for diverse dialogue generation.
\newblock In \emph{ICLR}.

\bibitem[{Wen et~al.(2023{\natexlab{b}})Wen, Li, Du, and Mou}]{wen-etal-2023-f}
Wen, Y.; Li, Z.; Du, W.; and Mou, L. 2023{\natexlab{b}}.
\newblock $f$-divergence minimization for sequence-level knowledge distillation.
\newblock In \emph{ACL}, 10817--10834.

\bibitem[{Wicks and Duh(2022)}]{wicks-duh-2022-effects}
Wicks, R.; and Duh, K. 2022.
\newblock The effects of language Token prefixing for multilingual machine translation.
\newblock In \emph{AACL-IJCNLP}, 148--153.

\bibitem[{Wolf et~al.(2019)}]{wolf2019huggingface}
Wolf, T.; et~al. 2019.
\newblock Huggingface's {T}ransformers: State-of-the-art natural language processing.
\newblock \emph{arXiv preprint arXiv:1910.03771}.

\bibitem[{Wolpert(1992)}]{WOLPERT1992241}
Wolpert, D.~H. 1992.
\newblock Stacked generalization.
\newblock \emph{Neural Networks}, 5(2): 241--259.

\bibitem[{Wu, Wu, and Huang(2021)}]{wu-etal-2021-one}
Wu, C.; Wu, F.; and Huang, Y. 2021.
\newblock One teacher is enough? {P}re-trained language model distillation from multiple teachers.
\newblock In \emph{ACL-IJCNLP Findings}, 4408--4413.

\bibitem[{Wu and Wang(2007)}]{wu-wang-2007-pivot}
Wu, H.; and Wang, H. 2007.
\newblock Pivot language approach for phrase-based statistical machine translation.
\newblock In \emph{ACL}, 856--863.

\bibitem[{Wu and Wang(2009)}]{wu-wang-2009-revisiting}
Wu, H.; and Wang, H. 2009.
\newblock Revisiting pivot language approach for machine translation.
\newblock In \emph{ACL-IJCNLP}, 154--162.

\bibitem[{Yang, Lv, and Chen(2023)}]{yang2023survey}
Yang, Y.; Lv, H.; and Chen, N. 2023.
\newblock A survey on ensemble learning under the era of deep learning.
\newblock \emph{Artificial Intelligence Review}, 56(6): 5545--5589.

\bibitem[{Zaremoodi, Buntine, and Haffari(2018)}]{zaremoodi-etal-2018-adaptive}
Zaremoodi, P.; Buntine, W.; and Haffari, G. 2018.
\newblock Adaptive knowledge sharing in multi-task learning: Improving low-resource neural machine translation.
\newblock In \emph{ACL}, 656--661.

\bibitem[{Zhang et~al.(2020)Zhang, Williams, Titov, and Sennrich}]{zhang-etal-2020-improving}
Zhang, B.; Williams, P.; Titov, I.; and Sennrich, R. 2020.
\newblock Improving massively multilingual neural machine translation and zero-shot translation.
\newblock In \emph{ACL}, 1628--1639.

\end{thebibliography}

\appendix

\newpage

\section{Beam Search} \label{app:vbs}
We show the standard beam search in Algorithm~\ref{alg:vbs} for a comparison with our proposed EBBS.
In general, beam search takes a scorer $p$ as the input and approximately finds the highest-scored sequence, by expanding its search tree with all the vocabulary (Lines~6--9) but only keeping the top-$Z$ partial candidates (Line~10) at each generation step.
Unlike EBBS, beam search is not specifically designed to work with multiple scorers, and we show in our main analysis that applying beam search with averaged probabilities of the ensemble components is not an ideal approach for ensemble decoding.

\setcounter{AlgoLine}{0}
\begin{algorithm}[t]
    \KwInput{
    $\mathbf x$: input sentence\, \\ \quad\quad\:\:\:\: $Z$: beam size \\
    \quad\quad\quad\: $p$: scorer
    }
    $H \gets \emptyset$ \Comment{candidate outputs} \\
    ${B}_0 \gets \{\langle \textsc{bos}, 1 \rangle \}$ \Comment{beam candidates} \\
    \For{$t = 1, 2, \cdots$}{
        $\triangleright$ core of beam search \\
        $B \gets \emptyset$ \\
        \For{$\langle \mathbf y_{1:t-1}, p' \rangle \in {B}_{t-1}$}{
            \For{$\mathrm y \in V$}{
                $p' \gets p' \cdot p(\mathrm y |\mathbf y_{1:t-1}, \mathbf x) $ \\
                $B.\operatorname{add}(\langle \mathbf y_{1:t-1} \oplus \mathrm y, p' \rangle)$
            }
        }
        ${B}_t \gets B.\operatorname{top}(Z)$\\
        $\triangleright$ check for termination \\
        \For{$\langle \mathbf y_{1:t}, p' \rangle \in {B}_t$}{
            \If{$\mathrm y_t = \textsc{eos}$}  {
                $H.\operatorname{add}(\langle \mathbf y, p' \rangle)$ \\
                \If{$|H| = Z$}{
                    \textbf{return} $H.\operatorname{top}(1)$
                }
            }
        }
    }
    \caption{Beam Search}
    \label{alg:vbs}
\end{algorithm}

\section{Experimental Details} \label{sec:app-datasets}

\textbf{Dataset details.} We evaluated our methods using IWSLT 2017~\cite{cettolo-etal-2017-overview} and Europarl v7~\cite{koehn-2005-europarl}. Table~\ref{tab:codes} provides a summary of the languages.

The IWSLT 2017 translation dataset features multilingual data derived from TED talks.
We followed previous work and used a standard split for zero-shot evaluation~\citep{DBLP:journals/corr/abs-1710-01025,liu-etal-2021-improving-zero}.
In particular, IWSLT contains English-centric training data for Italian, Dutch, and Romanian, while evaluation is performed in six zero-shot directions.
IWSLT is a relatively small dataset, which tests our method's ability to generalize from few languages.

\begin{table}[t]
\centering
\resizebox{0.6\columnwidth}{!}{
\begin{tabular}{|cl|cc|}
\hline
Code & \multicolumn{1}{c|}{Language} & IWSLT        & Europarl     \\ \hline
da   & Danish                        &              & $\checkmark$ \\
de   & German                        &              & $\checkmark$ \\
en   & English                       & $\checkmark$ & $\checkmark$ \\
es   & Spanish                       &              & $\checkmark$ \\
fi   & Finnish                       &              & $\checkmark$ \\
fr   & French                        &              & $\checkmark$ \\
it   & Italian                       & $\checkmark$ & $\checkmark$ \\
nl   & Dutch                         & $\checkmark$ & $\checkmark$ \\
pt   & Portugese                     &              & $\checkmark$ \\
ro   & Romanian                      & $\checkmark$ &              \\ \hline
\end{tabular}
}
\caption{The languages in the IWSLT and Europarl datasets.}
\label{tab:codes}
\end{table}

Europarl is a multilingual dataset crawled from the proceedings of the European Parliament.
We again followed previous work~\citep{liu-etal-2021-improving-zero} and evaluated our methods with a standard split for the zero-shot setting, containing English-centric data for eight languages with a total of 56 zero-shot evaluation directions. We adopted their non-overlapping setting: in the original corpus, a sentence may be translated into multiple languages, and the non-overlapping setup chooses only one target translation for each input. This prevents potential data-leaking problems.
Europarl contains more data and languages than IWSLT, which further tests our method's ability to generalize across multiple languages.

\textbf{Implementation details.} We directly adopted the neural architecture and hyperparameters in~\cite{liu-etal-2021-improving-zero}. In particular, we used 5- and 8-layer encoder--decoder models for IWSLT and Europarl, respectively. For both datasets, we had 512 hidden units and 8 attention heads. Our BLEU scores are based on SacreBLEU~\cite{post-2018-call} with the following specifications: \texttt{\seqsplit{BLEU+case.mixed+numrefs.1+smooth.exp+tok.13a+version.1.5.1}}.

In our presentation of beam search and the proposed EBBS, we describe the scorer as the multiplication of step-wise probabilities. In implementation, we used the sum of log-probabilities for numerical stability.
Moreover, our EBBS is built on top of the popular fairseq framework~\citep{ott-etal-2019-fairseq}, using their beam search implementation as the backbone.
Consequently, we inherit standard beam search implementation techniques such as length normalization and max length constraints, which are not detailed in our pseudocode.

\begin{figure}[t]
    \centering
    \includegraphics[width=0.7\columnwidth]{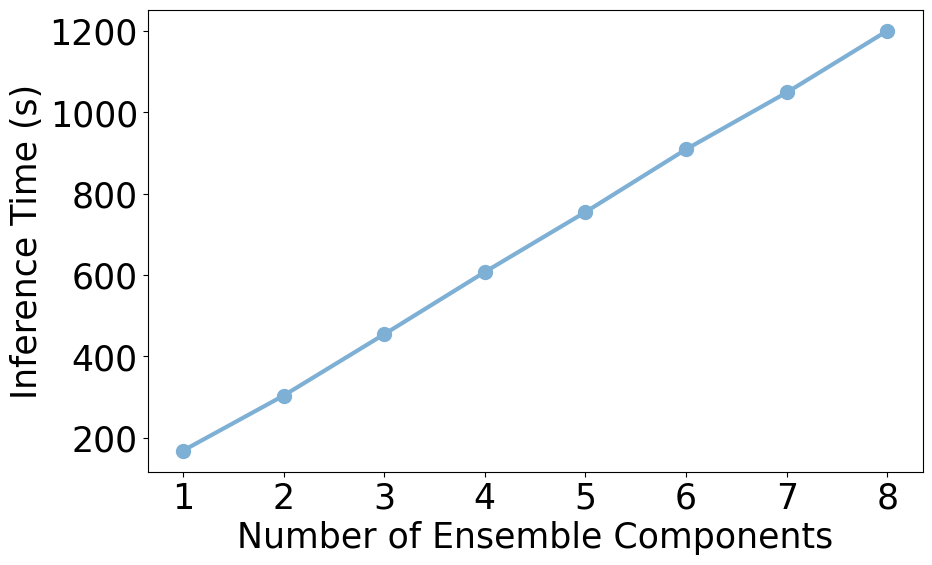}
    \caption{Inference time analysis on the test set of Italian-to-Dutch translation from Europarl. Experiments were conducted on an AMD EPYC 7313 CPU and an NVIDIA RTX A6000 GPU, with a batch size of 300 samples.}
    \label{fig:efficiency}
\end{figure}

\begin{table*}[t]
\centering
\resizebox{0.7\linewidth}{!}{
\begin{tabular}{|l|l|cccc|}
\hline
Dataset & Model & \multicolumn{1}{l}{BLEU$^\uparrow$} & \multicolumn{1}{l}{chrF2++$^\uparrow$} & \multicolumn{1}{l}{TER$^\downarrow$} & \multicolumn{1}{l|}{COMET$^\uparrow$} \\ \hline
\multirow{6}{*}{IWSLT} & Direct translation (our replication) & 17.29 & 42.57 & \textbf{72.46} & 0.7242 \\ \cline{2-6} 
 & Pivoting (en) & 16.19 & 42.79 & 80.39 & 0.7184 \\
 & Word-level averaging ensemble & 17.28 & 42.63 & 72.66 & 0.7237 \\
 & Word-level voting ensemble & 16.99 & 42.55 & 75.23 & 0.7162 \\
 & Sequence-level voting ensemble (MBR) & 16.72 & 42.50 & 76.12 & 0.7170 \\
 & EBBS (ours) & \textbf{18.24} & \textbf{43.66} & 73.50 & \textbf{0.7383} \\ \hline
\multirow{6}{*}{Europarl} & Direct translation (our replication) & 27.85 & 54.37 & 63.40 & 0.8426 \\ \cline{2-6} 
 & Pivoting (en) & 27.75 & 54.08 & 63.03 & 0.8480 \\
 & Word-level averaging ensemble & 27.78 & 54.45 & 63.80 & 0.8409 \\
 & Word-level voting ensemble & 27.62 & 54.05 & 63.11 & 0.8409 \\
 & Sequence-level voting ensemble (MBR) & 28.14 & 54.43 & 62.75 & 0.8452 \\
 & EBBS (ours) & \textbf{28.44} & \textbf{54.64} & \textbf{62.48} & \textbf{0.8488} \\ \hline
\end{tabular}
}
\caption{Average performance statistics across all zero-shot language pairs in IWSLT and Europarl.}
\label{tab:stats}
\end{table*}

\begin{table*}[]
\centering
\resizebox{0.8\linewidth}{!}{
\begin{tabular}{|cc|cccccccc|}
\hline
Lower & Upper & BLEU$^\uparrow$ & BLEU1$^\uparrow$ & BLEU2$^\uparrow$ & BLEU3$^\uparrow$ & BLEU4$^\uparrow$ & chrF2++$^\uparrow$ & TER$^\downarrow$ & COMET$^\uparrow$ \\ \hline
1 & 1 & 24.98 & 56.75 & 30.08 & 18.84 & 12.46 & 51.97 & 66.13 & 0.8255 \\
3 & 3 & 25.99 & 57.04 & 30.92 & 19.65 & 13.17 & 52.65 & \textbf{65.60} & 0.8326 \\
5 & 5 & \ul{26.10} & \textbf{57.07} & \textbf{31.00} & \ul{19.76} & 13.28 & \ul{52.75} & \ul{65.63} & 0.8340 \\
7 & 7 & \ul{26.10} & 57.06 & 30.98 & 19.74 & \ul{13.30} & \ul{52.75} & 65.69 & \ul{0.8346} \\
9 & 9 & \textbf{26.12} & 57.02 & 30.99 & \textbf{19.78} & \textbf{13.31} & \textbf{52.79} & 65.76 & \textbf{0.8352} \\ \hline \hline
5 & 1 & 25.06 & 56.70 & 30.10 & 18.88 & 12.49 & 52.07 & 66.20 & 0.8264 \\
5 & 2 & 25.63 & 56.90 & 30.56 & 19.29 & 12.87 & 52.48 & 65.92 & 0.8311 \\
5 & 3 & 25.94 & 56.93 & 30.87 & 19.61 & 13.14 & 52.69 & 65.79 & 0.8330 \\
5 & 4 & \ul{26.03} & \ul{57.02} & \ul{30.95} & \ul{19.69} & \ul{13.21} & \ul{52.74} & \ul{65.75} & \ul{0.8332} \\
5 & 5 & \textbf{26.10} & \textbf{57.07} & \textbf{31.00} & \textbf{19.76} & \textbf{13.28} & \textbf{52.75} & \textbf{65.63} & \textbf{0.8340} \\ \hline
\end{tabular}
}
\caption{Beam size analysis for Italian-to-Dutch translation on the Europarl dataset.}
\label{tab:beam}
\end{table*}

\section{Analysis of Inference Efficiency} \label{sec:app-inf}
We analyze the efficiency of our ensemble approach. As seen in Figure~\ref{fig:efficiency}, the inference scales almost linearly, which is reasonable as we need to perform inference for all the components. The trend shows that it is computationally feasible to build an ensemble of even more components.
Importantly, recall this linearly increasing inference time is mitigated by EBBS distillation (Row 8, Table~\ref{tab:main}), which is as fast as one component.

\section{Average Performance across Tasks}

In addition to the main results (Table~\ref{tab:main}), we further show the average performance across the 6 and 56 translation directions on IWSLT and Europarl using various metrics.
We omit EBBS distillation, because distilling across all directions requires significant computing resources.
As shown in Table~\ref{tab:stats}, EBBS consistently outperforms existing baselines, with the only exception for TER on the IWSLT dataset, where direct translation achieved a higher performance.
This is potentially because the TER score is based on edit distance, which is not robust to word reordering.
In general, results show EBBS consistently outperforms competing methods.

\section{Analysis of Beam Size} \label{app:beam}

We analyze the effect of different beam sizes on our EBBS algorithm.
First, we study the setting where the lower- and upper-level beam sizes are matched.
As seen in the top half of Table~\ref{tab:beam}, the performance tends to increase with a larger beam size and eventually plateaus at around five, which is consistent with the practice of standard beam search~\cite{meister-etal-2020-beam}.

Further, we analyze the setting where the upper- and lower-level beam sizes are not matched.
Generally, the upper-level beam size should not exceed the lower-level beam size, because otherwise the upper-level beam may not be fully filled by the ensemble components.
As shown in the bottom half of Table~\ref{tab:beam}, EBBS performs better with larger upper-level beam sizes.
This is understandable because a larger upper-level beam allows EBBS to explore more candidates in general.

Overall, our analysis shows that EBBS is robust and works well with a variety of beam sizes. Based on this experiment and efficiency considerations, we used a beam size of five for both upper- and lower-level beams in our main experiments.

\begin{table}[t]
\centering
\resizebox{0.8\columnwidth}{!}{
\begin{tabular}{|l|ll|cc|}
\hline
Dataset & \multicolumn{2}{c|}{Method} & BLEU & Entropy \\ \hline
\multirow{5}{*}{IWSLT} & \multicolumn{2}{l|}{EBBS} & \ul{19.52} & - \\ \cline{2-5} 
 & \multicolumn{1}{l|}{\multirow{4}{*}{\begin{tabular}[c]{@{}l@{}}Direct \\ translation\end{tabular}}} & No distillation & 17.46 & 2.46 \\
 & \multicolumn{1}{l|}{} & Direct distillation & 18.10 & 1.62 \\
 & \multicolumn{1}{l|}{} & Union distillation & 17.80 & 1.80 \\
 & \multicolumn{1}{l|}{} & EBBS distillation & \textbf{20.13} & 1.70 \\ \hline\hline
\multirow{5}{*}{Europarl} & \multicolumn{2}{l|}{EBBS} & \textbf{26.10} & - \\ \cline{2-5} 
 & \multicolumn{1}{l|}{\multirow{4}{*}{\begin{tabular}[c]{@{}l@{}}Direct \\ translation\end{tabular}}} & No distillation & 25.33 & 2.06 \\
 & \multicolumn{1}{l|}{} & Direct distillation & 25.44 & 1.44 \\
 & \multicolumn{1}{l|}{} & Union distillation & 25.53 & 1.59 \\
 & \multicolumn{1}{l|}{} & EBBS distillation & \ul{25.92} & 1.51 \\ \hline
\end{tabular}
}
\caption{Entropy of various distillation techniques on Italian-to-Dutch translation.}
\label{tab:entropy}
\end{table}

\begin{table*}[t]
\centering
\resizebox{0.85\linewidth}{!}{
\begin{tabular}{|l|cccccccc|}
\hline
Voting scheme & BLEU$^\uparrow$ & BLEU1$^\uparrow$ & BLEU2$^\uparrow$ & BLEU3$^\uparrow$ & BLEU4$^\uparrow$ & chrF2++$^\uparrow$ & TER$^\downarrow$ & COMET$^\uparrow$ \\ \hline
None (beam search) & 25.33 & 56.32 & 30.08 & 19.01 & 12.78 & 52.32 & 66.56 & 0.8276 \\ \hline
Total-sum & 25.27 & 56.67 & 30.27 & 19.07 & 12.65 & 52.19 & 66.12 & 0.8311 \\
Max & 25.81 & 56.89 & 30.76 & \ul{19.51} & \ul{13.09} & 52.46 & \ul{65.92} & 0.8300 \\
0/1 & \ul{25.84} & \ul{56.99} & \ul{30.78} & 19.49 & 13.05 & \ul{52.61} & 65.80 & \ul{0.8322} \\
Top-$Z$ sum (ours) & \textbf{26.10} & \textbf{57.07} & \textbf{31.00} & \textbf{19.76} & \textbf{13.28} & \textbf{52.75} & \textbf{65.63} & \textbf{0.8340} \\ \hline
\end{tabular}
}
\caption{Comparison of different ensemble variants, using Italian-to-Dutch translation in the Europarl dataset as the testbed.}
\label{tab:variants}
\end{table*}

\section{Entropy of Distilled Models} \label{app:disitllation-entropy}

We would like to understand why EBBS-based distillation largely outperforms other methods, such as union distillation (\S\ref{subsec:results}). Our hypothesis is that cross-entropy distillation loss with diverse samples may lead to an overly smooth distribution, which in turn would affect the model performance~\cite{wen-etal-2023-f,shayegh2023ensemble}. 

We show the average prediction entropy of our distilled models in Table~\ref{tab:entropy}. 
For some input $\mathbf x$ and generation step $t$, the prediction entropy is
\begin{align}\nonumber
     H = - \sum_{\mathrm y \in V}  & \;p(\mathrm y|\hat{\mathbf y}_{1:t-1}, \mathbf x) \log p(\mathrm y|\hat{\mathbf y}_{1:t-1}, \mathbf x) 
\end{align}
A large entropy generally indicates that the model is less certain, producing a more uniform prediction, whereas a low entropy indicates that the model is confident, producing a more peaked distribution.

As seen in Table~\ref{tab:entropy}, the model without distillation yields the highest entropy, suggesting that it is uncertain about zero-shot translation probably due to a lack of training signals.

Union distillation trains the model from the union of ensemble components' outputs. It reduces prediction entropy compared with no distillation, but due to the nature of cross-entropy loss, it remains the highest among all distillation variants. 
Direct distillation is based on direct translation only, reinforcing the model's current belief and thus producing the lowest entropy. On the contrary, our EBBS-based distillation achieves a moderate entropy on both datasets.

It should be emphasized that the entropy analysis merely shed light on how different distillation methods behave, but the entropy itself does not indicate the quality of a model. 
We quote BLEU scores from Table~\ref{tab:distill}, which has suggested that our EBBS-based distillation achieves similar or higher performance compared with EBBS, consistently outperforming other distillation methods.

\section{Analysis of Voting Methods in EBBS} \label{app:voting} In our EBBS algorithm, the lower-level beams are synchronized into a shared upper-level beam by voting.
Specifically, EBBS uses a mechanism of top-$Z$ sum voting, where we add the ensemble components' probabilities for each appearance of a candidate in the lower-level beam, shown in Eqn.~\eqref{eq:sum-vote}.
Here, we analyze a few alternative voting methods for EBBS.

If EBBS adopts total-sum voting, it still uses lower-level beams to find candidates, but adds all components' probabilities together. This is equivalent to applying the common averaging ensemble to the top-$Z$ candidates. However, it
differs from our approach, because in total-sum voting, a component will vote even if the candidate does not appear in its own lower-level beam; the probability after voting in Eqn.~(\ref{eq:sum-vote}) is substituted with $\frac1K\sum_k p_k(\mathbf y|\mathbf x)$.
As shown in Table~\ref{tab:variants}, EBBS with total-sum voting performs worse than direct translation, suggesting the importance of ignoring the components whose lower-level beam does not contain the candidate. This is analogous to nucleus sampling~\cite{holtzman2019curious}, where the long tail of a distribution is mainly noise and should be ignored. 

Other voting schemes that EBBS may use include 0/1 voting and max voting.
The former selects the candidates that appear most in the lower-level beams, disregarding the probability values (unless for ties);
the latter chooses the maximum probability across the lower-level beams, which gives preference to sequences through a maximization bias~\citep{NIPS2010_091d584f,Hasselt_Guez_Silver_2016}.
As seen, EBBS performs relatively well with both of these voting schemes, achieving a decent improvement over the baseline approach;
however, their performance is worse than our top-$Z$ sum voting.

Overall, the proposed bi-level beam search ensemble is effective with different voting schemes (except for the total-sum voting), and our top-$Z$ sum voting works the best among these variants. 

\begin{table*}[t]
\centering
\resizebox{\linewidth}{!}{
\begin{tabular}{|ll|}
\hline
\multicolumn{2}{|c|}{\textbf{IWSLT}} \\ \hline
\multicolumn{1}{|l|}{Input} & \begin{tabular}[c]{@{}l@{}}ho sempre creduto che trasformare la paura in divertimento sia il dono della creatività.\\ (\textit{I have always believed that turning fear into fun is the gift of creativity.}) \end{tabular} \\ [0pt] \hline
\multicolumn{1}{|l|}{Reference} & \begin{tabular}[c]{@{}l@{}}Ik heb altijd geloofd dat het omzetten van angst in plezier de gift is van creativiteit.\\ (\textit{I have always believed that turning fear into joy is the gift of creativity.}) \end{tabular} \\ \hline
\multicolumn{1}{|l|}{Direct translation} & \begin{tabular}[c]{@{}l@{}}Ik geloofde altijd dat het transformeren van angst in ontspanning de gift van creativiteit is\\ (\textit{I always believed that transforming anxiety into relaxation is the gift of creativity.})\end{tabular} \\ [0pt] \hline
\multicolumn{1}{|l|}{English-pivoting} & \begin{tabular}[c]{@{}l@{}}Omdat ik altijd geloofde om angst in plezier te transformeren, is het geschenk van creativiteit.\\ (\textit{Because I always believed to transform fear into pleasure is the gift of creativity.})\end{tabular} \\ [0pt] \hline
\multicolumn{1}{|l|}{Romanian-pivoting} & \begin{tabular}[c]{@{}l@{}}In feite hebben we altijd gedacht dat het transformeren van angst in divergentie de gift van creativiteit is.\\ (\textit{In fact, we have always thought that transforming fear into divergence is the gift of creativity}) \end{tabular} \\ [0pt] \hline
\multicolumn{1}{|l|}{EBBS} & \begin{tabular}[c]{@{}l@{}}Ik geloofde altijd dat het transformeren van angst in plezier de gift van creativiteit is.\\ (\textit{I always believed that transforming fear into pleasure is the gift of creativity.}) \end{tabular} \\ \hline\hline
\multicolumn{2}{|c|}{\textbf{Europarl}} \\ \hline
\multicolumn{1}{|l|}{Input} & \begin{tabular}[c]{@{}l@{}}si poteva avvertire una forte tensione.\\ (\textit{a strong tension could be felt.})\end{tabular} \\ [0pt] \hline
\multicolumn{1}{|l|}{Reference} & \begin{tabular}[c]{@{}l@{}}Er was veel spanning zichtbaar.\\ (\textit{There was a lot of tension visible.}) \end{tabular} \\ \hline
\multicolumn{1}{|l|}{Direct translation} & \begin{tabular}[c]{@{}l@{}}Er was grote spanning te ontgaan.\\ (\textit{There was great tension to be escaped.})\end{tabular} \\ [0pt] \hline
\multicolumn{1}{|l|}{English-pivoting} & \begin{tabular}[c]{@{}l@{}}Er zou veel spanningen kunnen zijn ontstaan.\\ (\textit{A lot of tensions could have arisen.}) \end{tabular} \\ [0pt] \hline
\multicolumn{1}{|l|}{Spanish-pivoting} & \begin{tabular}[c]{@{}l@{}}Mocht een sterke spanning kunnen worden aangekondigd.\\ (\textit{Should a strong tension can be announced.})\end{tabular} \\ [0pt] \hline
\multicolumn{1}{|l|}{EBBS} & \begin{tabular}[c]{@{}l@{}}Er was veel spanning geweest.\\ (\textit{There had been a lot of tension.}) \end{tabular} \\ \hline
\end{tabular}
}
\caption{Case studies, where the source language is Italian and the target is Dutch. We provide English interpretations in (\textit{italic}) for non-English text using Google Translate.}
\label{tab:cases}
\end{table*}

\section{Case Study} \label{app:case}
Table~\ref{tab:cases} shows examples of direct, pivot, and EBBS translations.
As seen, pivot and direction translations are prone to low-quality output, but EBBS enables them to correct each other's mistakes. 
In the first example, say, our EBBS generally follows the sentence structure of direct translation, where the Italian word ``divertimento'' (\textit{fun}) is mistranslated to the Dutch word ``ontspanning'' (\textit{relaxation}), but our EBBS corrects it to ``plezier'' (\textit{pleasure}), advocated by English pivoting and voted by all ensemble components. 

\end{document}